\documentclass[10pt]{article} 
\usepackage[preprint]{tmlr}

\usepackage{amsmath,amsfonts,bm}

\def\eqref#1{equation~\ref{#1}}

\def\1{\bm{1}}

\DeclareMathAlphabet{\mathsfit}{\encodingdefault}{\sfdefault}{m}{sl}
\SetMathAlphabet{\mathsfit}{bold}{\encodingdefault}{\sfdefault}{bx}{n}

\usepackage{amssymb}
\usepackage{graphicx}
\usepackage{booktabs}
\usepackage{xcolor}
\usepackage[colorlinks=true,allcolors=blue!50!black]{hyperref}
\usepackage{url}
\usepackage{caption}
\usepackage{subcaption}
\usepackage{tikz}
\usetikzlibrary{arrows.meta,positioning,fit,backgrounds}
\graphicspath{{figures/}}

\newcommand{\full}{\textsf{GoalDyn}}            
\newcommand{\obs}{\textsf{GoalDyn+Read}}        
\newcommand{\delk}{\textsf{GoalFree}}           
\newcommand{\anchors}{\textsf{NoGoal}}          
\newcommand{\textgoal}{\textsf{TextGoal}}       
\newcommand{\objtoken}{\textsf{ObjToken}}       
\newcommand{\geompred}{predicted-anchor accuracy}
\newcommand{\geomobs}{observed-anchor accuracy}
\newcommand{\ci}[2]{{\scriptsize[#1,#2]}}

\title{Grounding Spatial Relations in a Compact World Model:\\
Instruction Leakage and a Goal-Free Dynamics Fix}
\author{%
  Yufeng Wang$^{*1}$ \quad Lu Wei$^{*1}$ \quad Haibin Ling$^{\dagger 2}$\\[4pt]
  $^1$Stony Brook University \quad $^2$Westlake University\\[2pt]
  {\small $^*$Equal contribution \quad $^\dagger$Corresponding author}
}
\date{}

\begin{document}
\maketitle

\begin{abstract}
Compact world models that condition on a language goal promise to ground relations such as ``put the red block left of the blue block'' using a sparse set of explicit \emph{reference anchors}. We ask when such references actually ground a relation, and identify a trap: a goal-conditioned predictor reaches a striking $0.90$ relation-readout accuracy, yet this is \emph{instruction transcription}, not perception. Withholding the goal collapses it to chance ($0.90\!\to\!0.27$, three seeds) and a counterfactual instruction makes the predicted anchors follow the \emph{false} instruction $94.5\%$ of the time (true scene $2.3\%$; $N{=}256$). Tested across three settings and a within-task ablation, our central claim characterizes the confound: \textbf{instruction leakage occurs when the scored quantity is transcribable from the instruction (when the instruction names the answer) and is essentially independent of how predictive the non-instruction inputs are.} Our tabletop and the external BabyAI benchmark leak, whereas a Language-Table forward-dynamics world model whose instruction names \emph{referents} does not, until the instruction is augmented to name the direction; and degrading the action never increases leakage, the opposite of what predictor-competition predicts. The diagnosis prescribes the fix: keep the goal out of the dynamics (it belongs to the planner's cost) and supervise the \emph{read} path, recovering genuine, instruction-independent grounding ($0.88$, identical with and without the goal). The detection protocol and remedy apply to any goal-conditioned world model whose instruction names the scored quantity.
\end{abstract}

\section{Introduction}
A central promise of world models for embodied control is that they let an agent \emph{imagine} the consequences of its actions and plan against a goal \citep{worldmodels,planet,dreamer,tdmpc2,vjepa2}. For \emph{relational} goals such as ``put X to the left of Y'', the agent must represent not just appearance but the inter-object spatial relation. A popular and compact recipe pairs a joint-embedding predictive (JEPA) latent with a small set of explicit, metric \emph{reference anchors} and a language goal token \citep{lewm,causaljepa}: the anchors localize the referents, the latent carries the rest, and the goal conditions behavior. The intuition is that explicit coordinates give a relation a place to live that an entangled global latent does not.

We set out to measure this intuition rather than assert it: \emph{when, and why, do explicit references help a compact world model represent and act on a spatial relation, and when does a text-conditioned latent already suffice?} We build a controlled 2D relational tabletop, a diagnostic task in the spirit of CLEVR \citep{clevr}, in which referential ambiguity is a tunable knob (the number of duplicate distractors), train a ladder of models that vary the representation and the supervision, and evaluate them both as \emph{representations} (can a readout recover the relation?) and as \emph{controllers} (can a planner achieve the relation?).

The study turned on a confound that is worth stating up front because it is easy to fall into. A goal-conditioned model appears to ground relations extremely well: a geometric readout from its predicted anchors reaches $0.90$ accuracy and is, deceptively, robust to ambiguity. That number is almost entirely an artifact: because the instruction names the relation being scored, the predictor \emph{transcribes} the instruction rather than perceiving the scene. Two controls, withholding the goal and substituting a counterfactual one, make this precise; the apparent grounding evaporates under both. Generalizing the phenomenon into a falsifiable characterization is the paper's central claim: \textbf{instruction leakage in a goal-conditioned world model occurs when the scored quantity is \emph{transcribable from the instruction} (when the instruction names the answer) and is essentially independent of how predictive the non-instruction inputs (action, state) are.} We establish this across three settings: (1) our tabletop and the external BabyAI benchmark, where the instruction names the relation, both \emph{leak}, (2) whereas Language-Table, a forward-dynamics world model whose instruction names \emph{referents} rather than the answer, does \emph{not}. (3) A within-task action-ablation in which degrading the action \emph{never increases} leakage (the opposite of what competition with the action predicts) shows that the model transcribes even against a perfect action (\S\ref{sec:external}).

This diagnosis points directly at a fix. If the goal leaks through the dynamics, then the goal does not belong in the dynamics: a world model's transition should predict the consequences of \emph{actions}, and the goal should enter only through the planner's cost, where we use it anyway. We therefore train a variant with \emph{goal-free dynamics} (the predictor never sees the goal) and a supervised \emph{read} path for the anchors. This model grounds relations genuinely (its anchors encode the relation independent of the instruction) and, on the control side, confirms that goal-conditioning was actively hurting: removing it recovers control to the level of the simplest no-goal baseline. The remaining gap to an oracle is attributable to dynamics fidelity rather than to the representation.

\paragraph{Contributions.} Our contributions are summarized as follows:
\begin{enumerate}\itemsep2pt
\item \textbf{A characterization of instruction leakage.} We show, across three settings and a within-task ablation, that the confound is governed by \emph{transcribability}: it appears when the instruction names the scored quantity and is essentially independent of non-instruction predictor strength. Tabletop and BabyAI (instruction names the relation) leak; a Language-Table forward-dynamics world model (instruction names referents) does not, and augmenting its instruction to name the direction induces the leak; an action-ablation dose-response finds leakage never increases as the action is degraded, the opposite of predictor-competition (\S\ref{sec:external}).
\item \textbf{A diagnosis with a validated detection protocol.} A goal-conditioned model reaches $0.90$ relation-readout accuracy that is pure transcription: it collapses to chance ($0.90\!\to\!0.27$, 3 seeds) when the goal is withheld and follows a \emph{counterfactual} instruction $94.5\%$ of the time ($N{=}256$; Fig.~\ref{fig:leakage},~\ref{fig:counterfactual}). The goal-withheld and counterfactual probes (with a validated positive control: an engineered-leaky model fires them at $0.97$) cleanly separate perception from transcription, and must be run on the training render distribution.
\item \textbf{A fix, whose payoff is grounding.} Moving the goal out of the dynamics and supervising the read path recovers genuine, instruction-independent grounding ($0.88$, identical with and without the goal); no other variant does. Because the model leaks even against a perfect action, it is \emph{removing the goal from the dynamics}, not better features, that is required. On control, goal-conditioning monotonically hurts (no-goal $0.30$ $>$ goal-as-predictor $0.14$ $>$ goal-in-rollout $\approx$ floor; 3 seeds); the fix recovers control to the simplest no-goal baseline (a tie at $0.29$, not a gain), so its value is grounded \emph{perception}.
\item \textbf{Honest boundaries.} A sharp ambiguity boundary for the grounding (Fig.~\ref{fig:perception}) that does not transfer across ambiguity levels, and a residual control gap (all variants $\ll$ oracle) attributable to dynamics fidelity, not the representation.
\end{enumerate}

\section{Related work}

\paragraph{Compact JEPA world models.}
Our backbone follows the LeWM family \citep{lewm} and instantiates the joint-embedding predictive architecture, or JEPA, framework \citep{jepa,ijepa}. In this setting, a compact latent representation is rolled forward by an action-conditioned predictor and used for goal-directed planning at test time. Related models predict dense visual features \citep{dinowm} or scale JEPA-style prediction to video \citep{vjepa2}, but relational information, when present, is typically stored implicitly inside an opaque latent state. Our work asks whether adding an explicit anchor channel can make such relations more grounded. We find that this depends critically on how the goal enters the model: anchors help only when the model is forced to infer the relation from perception rather than copy it from the goal.

\paragraph{Structured, object-centric, and point-based world models.}
A broad line of work adds spatial structure to world models through object-centric representations or point-based dynamics. Object-centric JEPAs \citep{causaljepa,dyno,sold,slotformer}, building on slot-based object representations \citep{slotattn}, represent scenes as collections of object-like latents, while point and particle dynamics models \citep{particleformer,terra,pri4r} model spatial change through explicit points or particles. The closest thread is object-centric JEPA \citep{causaljepa}: because object latents can already encode referential structure, explicit coordinates must provide information that the latent representation does not reliably capture on its own. Factored latent-action world models \citep{flam} also decompose the scene, but they factor the \emph{action} space through per-factor latent actions, without a JEPA joint-embedding objective or a text goal. In contrast, our factorization is \emph{representational}: we use a latent state $z$ together with explicit metric anchors, trained under a JEPA objective with a goal. We therefore do not claim a new architecture; instead, we characterize a failure mode, instruction leakage through goal-conditioned dynamics, that can arise in many structured world models whenever the goal names the relation being evaluated. We also provide a remedy that these models can adopt. Prior work on privileged point supervision discarded at inference \citep{pri4r} is complementary to our \emph{persistent} supervised read path, and known identity-drift failures in object-centric methods further motivate explicit supervision \citep{tcslots}.

\paragraph{Language-conditioned planning and verify-then-edit.}
Conditioning the transition model on language is an established design choice, not a strawman. For example, \citet{thinkjepa} inject language features into the JEPA predictor using FiLM \citep{film}, and instruction-conditioned video models, world models, and language-conditioned manipulation policies \citep{cliport,vima} commonly feed the instruction into the dynamics. Other approaches keep language out of the transition: some condition the JEPA \emph{cost} instead of the predictor \citep{pijepa}, some use language as a pretext signal \citep{vlajepa}, and others re-score actions or plans against a language goal at the control layer \citep{vlwm,vlmpc}. Related image-domain work verifies and edits generations according to spatial relations \citep{uig}. Our point is that feeding the goal into the transition can backfire in a specific but important setting: when the instruction directly names the relation being evaluated. In that case, the predictor can copy the goal information, allowing a relation readout to appear correct without requiring the model to perceive or ground the relation in the scene. Placing the goal in the cost rather than the dynamics follows the standard planning setup and, as we show, avoids this leakage. Thus, our contribution is not to claim that cost-conditioning is entirely new, but to show why goal-conditioned dynamics are unsafe for relation-naming instructions and to provide evidence for a safer alternative.

\paragraph{Shortcut learning.}
Instruction transcription is a form of shortcut learning \citep{shortcut}: the model exploits a spurious but predictive signal, the instruction text, instead of the intended signal, the scene. This is analogous to hypothesis-only baselines in natural language inference, where the hypothesis alone can often solve the task \citep{hyponli,annotartifacts}, and to language priors in visual question answering, where the question alone can predict the answer without using the image \citep{vqav2,vqaprior}. Such shortcuts are also a manifestation of simplicity bias \citep{simplicitybias}, where models prefer easier predictive signals even when those signals do not reflect the intended reasoning process. Our contribution is to identify a concrete and quantified instance of this problem in goal-conditioned world models, propose a detection protocol for exposing it, and show an architectural remedy that prevents relation readouts from being satisfied through instruction leakage alone.

\section{Task and model}
\label{sec:task}

\paragraph{Relational tabletop.} We procedurally generate scenes of $N{=}4$ colored shapes on a board. One object is the \emph{target} (slot~0) and one the \emph{anchor} (slot~1); the relation $r\in\{\textsf{left},\textsf{right},\textsf{above},\textsf{below}\}$ is the dominant axis of the target relative to the anchor. Referential \emph{ambiguity} $a$ is the number of duplicate (color,\,shape) distractors ($a\in\{0,\dots,4\}$), so language alone cannot always pick the referent. Each episode renders an image sequence; the target is pushed by a 2D action and the kinematics are a deterministic stepper, so the relation can be changed by acting. A language goal is tokenized from a template (``move the \textit{$\langle$target$\rangle$} to the \textit{$\langle r\rangle$} of the \textit{$\langle$anchor$\rangle$}''). \emph{This template names $r$}, the property that creates the leakage we study (\S\ref{sec:leak}). Four-way chance is $0.25$.

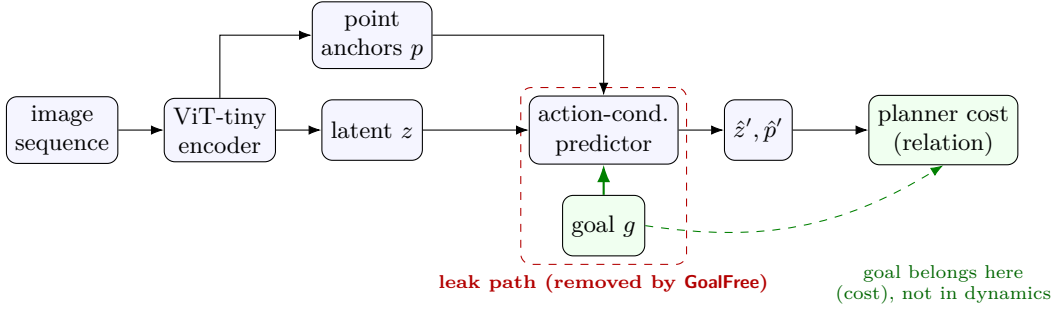
\begin{figure}[t]\centering
\begin{tikzpicture}[font=\small,node distance=6mm,
  box/.style={draw,rounded corners,minimum height=8mm,align=center,fill=blue!4},
  gbox/.style={draw,rounded corners,minimum height=8mm,align=center,fill=green!6}]
  \node[box](img){image\\sequence};
  \node[box,right=of img](enc){ViT-tiny\\encoder};
  \node[box,right=of enc](z){latent $z$};
  \node[box,above=4mm of z](p){point\\anchors $p$};
  \node[box,right=14mm of z](pred){action-cond.\\predictor};
  \node[gbox,below=4mm of pred](goal){goal $g$};
  \node[box,right=of pred](next){$\hat z',\hat p'$};
  \node[gbox,right=10mm of next](cost){planner cost\\(relation)};
  \draw[-{Latex}] (img)--(enc); \draw[-{Latex}](enc)--(z); \draw[-{Latex}](enc)|-(p);
  \draw[-{Latex}](z)--(pred); \draw[-{Latex}](p)-|(pred);
  \draw[-{Latex}](pred)--(next);
  \draw[-{Latex},green!50!black,thick](goal)--(pred) node[midway,right,font=\scriptsize]{};
  \draw[-{Latex}](next)--(cost);
  \begin{scope}[on background layer]
    \node[draw,dashed,red!70!black,rounded corners,fit=(goal)(pred),inner sep=3pt,label=below:{\scriptsize\color{red!70!black}\textbf{leak path (removed by \delk)}}]{};
  \end{scope}
  \node[align=center,below=12mm of cost,font=\scriptsize,green!40!black]{goal belongs here\\(cost), not in dynamics};
  \draw[-{Latex},green!50!black,dashed] (goal.east) to[bend right=20] (cost.south);
\end{tikzpicture}
\caption{\textbf{PrismWM and the leak path.} State factors into a JEPA latent $z$, sparse metric point anchors $p$, and a goal $g$. In the goal-conditioned model the goal enters the \emph{dynamics} (red), letting the predictor transcribe the instruction into anchors. Our fix (\delk) deletes that edge: the goal enters only the planner cost (green), where it is used for control anyway.}
\label{fig:arch}
\end{figure}

\paragraph{PrismWM.} The model (Fig.~\ref{fig:arch}) factors state into a global JEPA latent $z$, a sparse set of metric point anchors $p$ (read from patch features), and a goal embedding $g$. A ViT-tiny encoder produces $z$ and features. A referential module \emph{reads} anchors from features, \emph{conditions} the latent on them, and \emph{decodes} predicted anchors from the predicted latent. An action-conditioned transformer predictor rolls $(z,p)$ forward. Training minimizes next-latent prediction, a JEPA regularizer, an anchor (primitive) loss, and a goal loss. The \emph{referential interface} is pluggable (point anchors here; object-token latents as a baseline), so the representation is a measured choice, not an asserted one.

\paragraph{The ladder.} We vary: anchors versus none (\anchors{} versus the text and latent baselines); goal versus no goal; supervising only the predicted anchors (\full) versus also the \emph{observed} (read) anchors (\obs); and goal-free dynamics, which is our our fix (named \delk, \S\ref{sec:fix}). We also train \emph{learned} (unsupervised) anchors, but our geometric readout cannot fairly score them (\S\ref{sec:limits}), so we report them only as a flagged negative rather than as an interpreted axis. All models are trained per ambiguity level.

\section{Evaluation protocol}
\label{sec:eval}
We separate \emph{representation} from \emph{control}, and within representation we separate \emph{perception} from \emph{transcription}.

\paragraph{Representation readouts.} From a model's anchors we compute the \geomobs{} (relation read geometrically from the \emph{observed} anchors of the current frame) and the \geompred{} (from the anchors after a one-step action-conditioned rollout). We also fit a linear probe on the global latent $z$ (within-distribution stratified split, balanced accuracy). All numbers use $n{=}512$ episodes and $1000$-sample bootstrap $95\%$ intervals.

\paragraph{Leakage controls (the crux).} Because the goal template names $r$, a goal-conditioned readout is confounded. We add two controls: \emph{(i) goal withheld}: zero the goal tokens and recompute the readout on the same in-distribution frames; and \emph{(ii) counterfactual goal}: feed a goal naming a \emph{different} relation and measure whether the anchors follow the (false) goal or the true scene. A genuine perceptual readout is invariant to (i) and follows the true scene under (ii). We verified that the encode and read path never uses the goal (its readout is identical with and without the goal for every model), so the \geomobs{} is a clean perception measure; the predict path is where transcription occurs.

\paragraph{Render distribution.} Leakage-controlled readouts must run on the \emph{training} render distribution. A quick variant on freshly generated scenes under-counted a render-sensitive supervised encoder ($0.35$ versus the in-distribution $0.84$). We use dataset frames throughout and report the fresh-scene probe only for the render-invariant counterfactual.

\paragraph{Control.} We turn the tabletop into a task: generate a scene whose goal relation \emph{differs} from the current one, and plan an $8$-step sequence of pushes by sampling-based model-predictive control (the cross-entropy method \citep{cem}, as in model-based control with learned dynamics \citep{pets}) that, at each step, re-encodes the real scene, rolls the model's anchors forward under candidate actions, and scores a soft relation-satisfaction cost. The first action is executed in the kinematic environment and the loop repeats. Success is the \emph{true} relation achieved. We report the model against a \emph{random} policy (floor) and an \emph{oracle} planner that uses the true dynamics (ceiling). For the goal-conditioned models the planner uses the model as a pure forward predictor (goal applied externally in the cost); a goal-in-rollout ablation quantifies how leakage destroys the control signal.

\section{Results}

\subsection{Apparent grounding is instruction transcription}
\label{sec:leak}
\begin{figure}[t]
\centering
\begin{subfigure}{0.60\textwidth}\includegraphics[width=\linewidth]{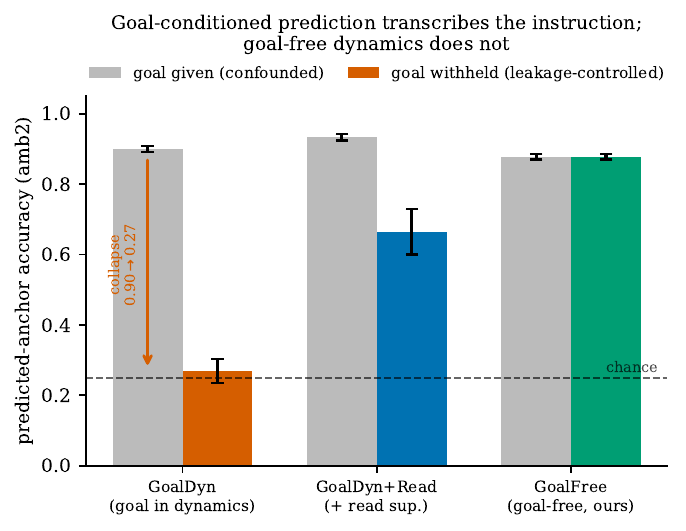}
\caption{}\label{fig:leakage}\end{subfigure}\hfill
\begin{subfigure}{0.37\textwidth}\includegraphics[width=\linewidth]{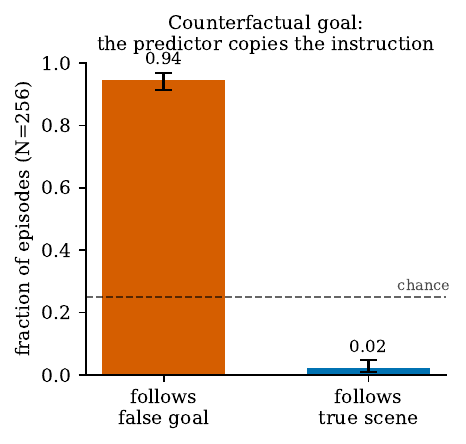}
\caption{}\label{fig:counterfactual}\end{subfigure}
\caption{\textbf{Instruction transcription.} (a) Predicted-anchor relation accuracy at medium ambiguity with the goal given (grey) versus withheld (color). The goal-conditioned model collapses to chance when the goal is withheld (leakage); adding observed-supervision partly de-leaks the predictor; goal-free dynamics (\delk, ours) is unchanged because it never uses the goal. (b) Under a counterfactual goal, \full's predicted anchors follow the false instruction, not the scene. Error bars in (a) are mean$\pm$s.d.\ over three seeds; (b) is a within-model probe with $95\%$ bootstrap intervals; the dashed line is four-way chance.}
\end{figure}

\begin{figure}[t]\centering
\includegraphics[width=0.92\textwidth]{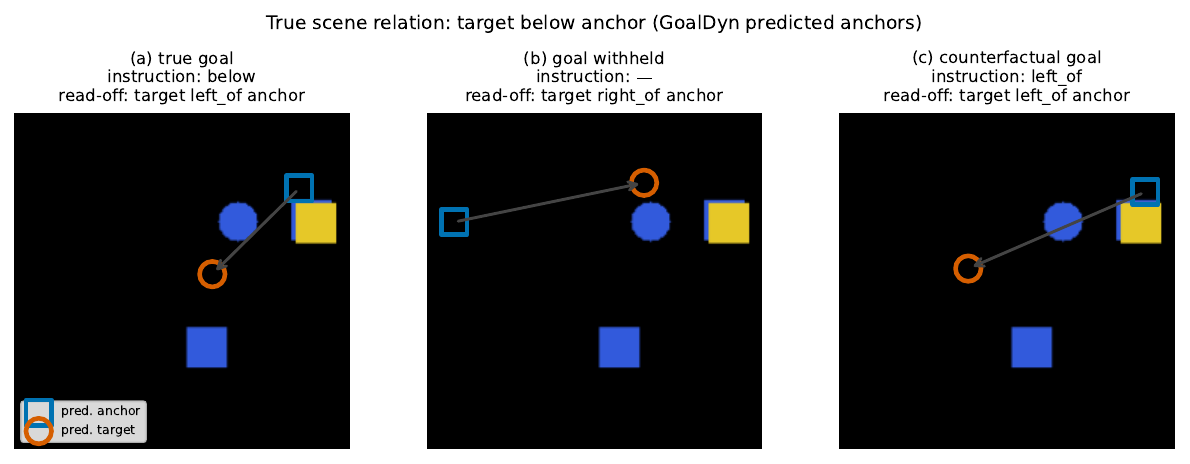}
\caption{\textbf{Transcription, visualized.} One tabletop scene (true relation: target \emph{below} anchor) with \full's \emph{predicted} anchors overlaid (target $\circ$, anchor $\square$) under three instructions. (a) True goal: the predicted target satisfies the named relation. (b) Goal withheld: the placement is uninformative. (c) Counterfactual goal (\emph{left of}): the predicted target moves to satisfy the \emph{false} instruction, not the scene; the model draws the answer from the instruction rather than perceiving it. This is a representative example, quantified in Fig.~\ref{fig:leakage},~\ref{fig:counterfactual}.}
\label{fig:qual}
\end{figure}

A goal-conditioned model with supervised anchors appears to ground relations almost perfectly: its \geompred{} is $\approx\!0.90$ and, deceptively, flat across ambiguity (Table~\ref{tab:boundary}, \full{} column). The leakage controls dismantle this. Withholding the goal collapses \full{}'s \geompred{} from $0.895$ to $0.262$ at medium ambiguity, i.e.\ chance, with non-overlapping intervals (Fig.~\ref{fig:leakage}). The collapse is stable across three training seeds ($0.900\pm0.009$ given $\to$ $0.270\pm0.034$ withheld; mean$\pm$s.d.). Under a counterfactual goal, \full{}'s predicted anchors follow the \emph{false} instruction $94.5\%$ \ci{0.914}{0.969} of the time and the true scene only $2.3\%$ \ci{0.008}{0.047} ($N{=}256$; Fig.~\ref{fig:counterfactual}), as visualized in Fig.~\ref{fig:qual}. The predicted-anchor ``grounding'' is the model copying the instruction into coordinates; ambiguity-robustness is trivial because the answer is in the prompt. The same root cause shows up in control: feeding the goal into the planner's rollout drops \full{}'s control success to the random floor ($0.047$ at $a{=}2$, $\le 0.05$ across ambiguities).

\subsection{Where genuine perception lives, and where it breaks}
\label{sec:perception}
\begin{figure}[t]\centering
\includegraphics[width=0.7\textwidth]{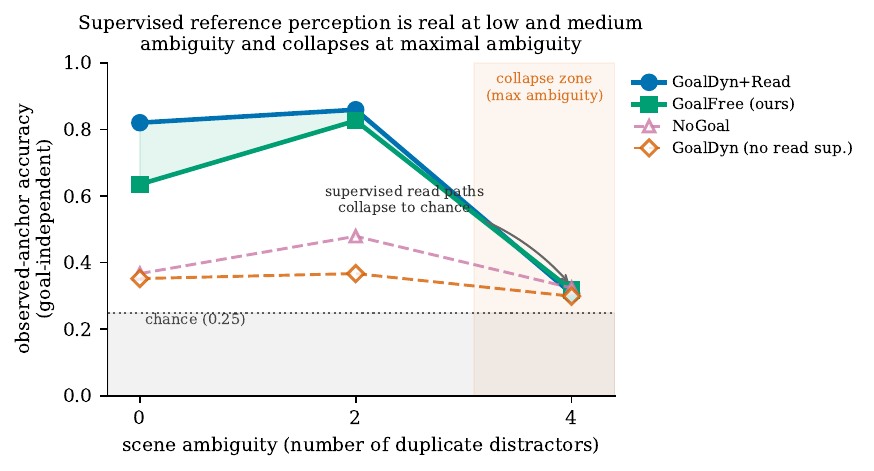}
\caption{\textbf{The ambiguity boundary of perception.} Goal-independent observed-anchor accuracy versus scene ambiguity. Supervised read paths (\obs, \delk) ground the relation at low and medium ambiguity and collapse at $a{=}4$; without observed-supervision (\full) or without anchors the read path is near chance throughout.}
\label{fig:perception}
\end{figure}

Genuine, goal-independent perception does exist in the \emph{observed} (read) anchors, when they are supervised. With observed-anchor supervision, \obs{} reads the true relation at $0.82$ and $0.86$ at ambiguity $0$ and $2$. This is verified leakage-free: identical with the true goal and with the goal zeroed on the same frames (Table~\ref{tab:boundary}; both the \geomobs{} and the latent probe are goal-invariant for all models). The relation is also \emph{partly} linearly decodable from the global latent (latent probe $0.54$--$0.60$ for the supervised-read models), but the explicit anchor channel carries it more cleanly (the \geomobs{} exceeds the latent probe throughout, e.g.\ $0.83$ versus $0.54$ for \delk{} at amb2), a quantitative form of the explicit-reference premise. Learned (\emph{unsupervised}) anchors score at chance on the geometric readout ($\approx\!0.25$). We \emph{flag} rather than interpret this, because the \geomobs{} assumes the supervised slot convention and is not a fair readout for free anchors (\S\ref{sec:limits}). Without observed-supervision the read path is weak (\full: $\approx\!0.37$). The grounding also has a sharp boundary: at maximal ambiguity ($a{=}4$, all distractors duplicated) it collapses to chance ($0.31$). This is the boundary between where references help and where they do not that the study set out to find (Fig.~\ref{fig:perception}). Moreover, the grounding does \emph{not transfer} across ambiguity: an $a{=}2$ model reads $0.86$ at $a{=}2$ but falls to chance ($0.26$--$0.33$) at every other test ambiguity, including the \emph{easier} $a{=}0$ (verified clean: the $a{=}0$ data yields $0.82$ for the $a{=}0$-\emph{trained} model). The learned grounding is thus narrowly tuned to its training ambiguity, not a general relational perceiver.

\begin{table}[t]\centering\small
\caption{\textbf{Leakage-controlled boundary sweep} ($n{=}512$, $95\%$ bootstrap intervals). The latent probe and observed-anchor accuracy are goal-invariant (identical with the true and withheld goal) and measure genuine perception; for supervised-read models the anchor channel carries the relation better than the latent (e.g.\ \delk{} amb2: $0.83$ vs.\ $0.54$). Predicted-anchor accuracy is shown with the goal given vs.\ withheld; the gap is the leakage. \textbf{Bold} in the withheld column marks the key evidence per model: for \full{}, the collapse to chance (leakage); for \delk{}, the match with the goal-given value (no leakage). Bold in the observed-anchor column marks genuine supervised perception (\obs). Chance $= 0.25$; ``n/a'' $=$ no anchor channel. Values are seed~0; three-seed mean$\pm$s.d.\ for the headline rows is in \S\ref{sec:fix}.}

\label{tab:boundary}
\resizebox{\textwidth}{!}{%
\begin{tabular}{llcccc}
\toprule
model & amb & latent probe & obs.\ acc.\ & pred.\ acc.\ (goal given) & pred.\ acc.\ (withheld)\\
\midrule
\full{} (goal in dynamics)  & 0 & 0.331 & 0.352 & 0.916 & \textbf{0.314}\\
                            & 2 & 0.290 & 0.367 & 0.895 & \textbf{0.262}\\
                            & 4 & 0.324 & 0.299 & 0.908 & \textbf{0.301}\\
\addlinespace
\obs{} (+ read sup.)        & 0 & 0.598 & \textbf{0.820} & 0.910 & 0.639\\
                            & 2 & 0.564 & \textbf{0.859} & 0.934 & 0.621\\
                            & 4 & 0.318 & 0.307 & 0.854 & 0.314\\
\addlinespace
\delk{} (goal-free, ours)   & 0 & 0.456 & 0.635 & 0.678 & \textbf{0.678}\\
                            & 2 & 0.543 & 0.826 & 0.869 & \textbf{0.869}\\
                            & 4 & 0.292 & 0.318 & 0.422 & \textbf{0.422}\\
\addlinespace
\anchors{} (no goal)        & 2 & 0.335 & 0.479 & 0.463 & 0.516\\
\textgoal{} (goal, no anchors) & 2 & 0.284 & $\approx$0.25 & $\approx$0.25 & $\approx$0.25\\
\objtoken{} (object latents)  & 2 & 0.303 & n/a & n/a & n/a\\
\bottomrule
\end{tabular}}
\end{table}

\subsection{Goal-free dynamics recovers grounding; goal-conditioning hurts control}
\label{sec:fix}
The diagnosis prescribes the fix: remove the goal from the dynamics. The proposed \delk{} trains the predictor without the goal (it is supplied only to the planner cost) and supervises the read path.

\emph{The payoff is genuine grounding.} \delk's \geompred{} is identical with and without the goal (single-seed $0.68$, $0.87$, $0.42$ for $a{=}0,2,4$): by construction there is nothing to leak. It is high where perception is available ($0.87$ at $a{=}2$ single-seed, $0.88$ averaged over three seeds), versus the leaked $0.90\!\to\!0.27$. No other variant gives instruction-independent predicted-anchor grounding; this is the fix's unique contribution.

\emph{For control, the clean result is that goal-conditioning hurts, not that our fix wins.} We are deliberate here. The evidence is a monotone gradient in how much the goal drives the dynamics, and across three training seeds (mean$\pm$s.d.) it is consistent: \anchors{} (no goal) $0.30\pm0.02$ and \delk{} (goal-free) $0.29\pm0.03$ both exceed \full{} (goal as predictor, cost external) $0.14\pm0.02$, which in turn exceeds \full{} with the goal in the rollout $0.03\pm0.02$ (at or below the random floor $0.055$). The ordering holds in \emph{every} seed: \delk{} $>$ \full{} in all three ($0.31, 0.31, 0.27$ versus $0.13, 0.12, 0.16$), as does \anchors{} $>$ \full{}, so the single-seed marginal interval we previously hedged is no longer load-bearing. But removing the goal recovers control \emph{only to the level of the simplest no-goal baseline}: \delk{} $0.29\pm0.03$ \emph{ties} \anchors{} $0.30\pm0.02$ (a wash). We therefore do not claim a control gain from the fix over the natural baseline; the fix's value is the supervised \emph{perception} \anchors{} lacks (Table~\ref{tab:boundary}), while its control merely confirms that goal-conditioning was the thing dragging \full{} down (\delk{} also exceeds \obs{}, $0.29\pm0.03$ versus $0.22\pm0.03$, in every seed: goal-in-dynamics hurts even with the read path supervised). All variants remain well below the oracle ($0.71$): the residual gap is dynamics \emph{fidelity} (a one-step-trained predictor whose multi-step rollouts drift, with under-estimated displacement), not the explicit-reference representation.

\begin{table}[t]\centering\small
\caption{\textbf{Closed-loop control success} ($n{=}128$, $95\%$ bootstrap intervals). Ordered by goal-in-dynamics: control falls monotonically as the goal gains sway over the dynamics; the goal-free model (\delk) \emph{ties} the simplest no-goal baseline (\anchors), and all variants are far below the oracle (a dynamics-fidelity bottleneck). amb2 intervals: \anchors{} $[0.242,0.391]$, \delk{} $[0.234,0.383]$, \full{} $[0.078,0.195]$, \full{}+goal-in-rollout $[0.016,0.086]$. Values are seed~0; the three-seed mean$\pm$s.d.\ underlies the \S\ref{sec:fix} headlines and the abstract.}
\label{tab:control}
\resizebox{.8\textwidth}{!}{%
\begin{tabular}{lccc|cc}
\toprule
model & amb0 & amb2 & amb4 & random & oracle\\
\midrule
\anchors{} (no goal) & 0.250 & 0.312 & 0.258 & & \\
\delk{} (goal-free, ours) & 0.258 & \textbf{0.312} & 0.219 & 0.05--0.06 & 0.71--0.79\\
\obs{} (goal in dynamics) & 0.250 & 0.211 & 0.172 & & \\
\full{} (goal as predictor) & 0.164 & 0.133 & 0.102 & & \\
\full{} + goal in rollout & 0.039 & 0.047 & 0.023 & & \\
\bottomrule
\end{tabular}}
\end{table}

\begin{figure}[t]\centering
\includegraphics[width=0.62\textwidth]{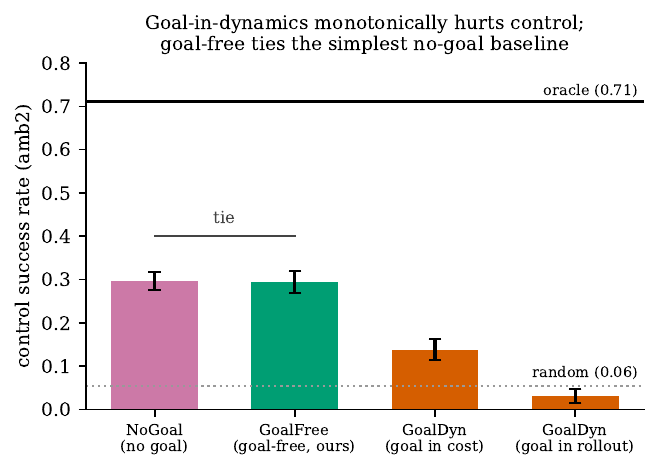}
\caption{\textbf{Goal-involvement gradient (medium ambiguity).} Control falls monotonically as the goal gains sway over the dynamics (no goal $\to$ goal as predictor $\to$ goal in rollout); the goal-free model (\delk) \emph{ties} the simplest no-goal baseline (\anchors), so the fix's payoff is grounded perception (Fig.~\ref{fig:perception}, Table~\ref{tab:boundary}), not a control gain. The gap to the oracle ceiling is dynamics fidelity. Bars are mean$\pm$s.d.\ over three seeds; the ordering holds in every seed.}
\label{fig:control}
\end{figure}

\subsection{When does the leakage occur? Three settings and a dose-response}
\label{sec:external}
We now establish the central claim: leakage is governed by \emph{transcribability} (whether the instruction names the scored quantity) and is essentially independent of how predictive the non-instruction inputs are. The counterfactual probe has a \emph{validated positive control} throughout: an engineered-leaky model fires it at $0.97$ (versus $0.03$ for a no-goal baseline), so a ``no leak'' reading is a real measurement, not a dead instrument.

\paragraph{BabyAI (external validation: relation-naming instructions cause leakage).}
We further validate this failure mode on \textbf{BabyAI} \citep{babyai}, specifically the GoToLocal level of MiniGrid. This setting differs from our generator in environment, modality, and relation type: it uses a symbolic gridworld, and the evaluated relation is the egocentric \emph{left}/\emph{right} relation, which is explicitly named in BabyAI instructions such as ``\dots\ on your left.'' A goal-conditioned readout that FiLMs the instruction into the predictor achieves a score of $1.00$ under the true instruction, but also follows a \emph{counterfactual} instruction with score $1.00$, while the true scene-consistent score is $0.00$. This occurs even though the perceptual problem itself is easy: an observation-only model reaches $0.86$. Thus, the failure is not due to insufficient perceptual information, but to a shortcut introduced by conditioning the dynamics on a relation-naming instruction. The same leakage therefore reproduces outside our own data generator.

\paragraph{Language-Table (external; instruction names \emph{referents} $\Rightarrow$ no leak).} We escalate to a goal-conditioned \emph{forward-dynamics} world model on Google's Language-Table \citep{langtable} (2D block pushing; $46{,}426$ goal-directed oracle transitions). Here the instruction names \emph{referents} (``push the blue cube towards the red moon''), so the scored quantity (the direction the target should move) is \emph{not} transcribable without the block positions. The same validated probe reports \textbf{no leak} (counterfactual cos $\approx 0$); a no-goal baseline matches the goal model's next-state MSE ($5.08\!\times\!10^{-6}$ versus $5.06\!\times\!10^{-6}$), so the goal is not needed for dynamics. Critically, \emph{augmenting} the instruction with a token that names the goal direction (making the answer transcribable) flips the counterfactual from $\approx0$ to $0.17$ (capped by the small per-step motion): the leak appears exactly when the instruction names the answer.

\paragraph{Dose-response: leakage is independent of non-instruction predictor strength.} A natural alternative hypothesis is that the model leaks only when the non-instruction inputs (here, the action) are weak predictors. We refute this with a within-task ablation that scales the action input ($\alpha:1\!\to\!0$), holding everything else fixed (Table~\ref{tab:dose}): \emph{degrading the action never increases leakage}, the opposite of what predictor-competition predicts. With a transcribable instruction and meaningful motion it stays high even against a \emph{perfect} action (synthetic, $0.975\!\to\!0.986$); without transcription it stays at chance (referent-only, $\approx0$); and in the $+$direction regime it stays weak and even \emph{declines} ($0.174\!\to\!0.032$); in no case does weakening the action \emph{induce} leakage. The synthetic is the load-bearing positive: its predicted displacement magnitude is robust ($\approx0.13$, Table~\ref{tab:dose}, last column), so the cosine is a strong probe there; the Language-Table regimes have $\approx10\times$ smaller motion ($\approx0.011$), making the cosine low-signal, so we read Language-Table only as ``transcription off $\Rightarrow$ no leak, augmenting it on $\Rightarrow$ a weak induced leak'' and do not over-interpret the $+$direction within-regime trend.\footnote{We had hypothesized the $+$direction decline was a vanishing-motion floor; the magnitude column refutes that (it is roughly constant in $\alpha$, $\approx0.011$), so we do not attribute the decline to a floor, nor to competition (which predicts the \emph{opposite}, an increase as $\alpha\!\to\!0$). The regime is simply low-signal.}

\begin{table}[t]\centering\small
\caption{\textbf{Action-ablation dose-response} (counterfactual cosine; positive control $0.97$; last column is the mean predicted-displacement magnitude under zero action). Degrading the action never \emph{increases} leakage in any regime, the opposite of predictor-competition; leakage is governed by whether the instruction names the answer. The synthetic carries the positive (robust motion $\approx0.13$); the Language-Table regimes have $\approx10\times$ smaller motion, so their cosine is low-signal.}
\label{tab:dose}
\begin{tabular}{lcccc}
\toprule
regime (counterfactual cosine) & $\alpha{=}1.0$ & $\alpha{=}0.5$ & $\alpha{=}0.0$ & disp.\ mag ($\alpha{:}1{\to}0$)\\
\midrule
synthetic (transcription available) & 0.975 & 0.984 & 0.986 & $0.12\!\to\!0.13$\\
Language-Table, referent-only (no transcription) & $-0.00$ & $-0.01$ & $-0.08$ & $0.009\!\to\!0.010$\\
Language-Table, $+$direction token (transcription on) & 0.174 & 0.161 & 0.032 & $0.011\!\to\!0.011$\\
\bottomrule
\end{tabular}
\end{table}

\paragraph{Which deployed designs are exposed?} The characterization predicts a structural split, which a survey of recent goal-conditioned models bears out: designs that place the goal in the planner's \emph{cost} or re-score rollouts against it (\citealp{vlwm,vlmpc,pijepa}) are \emph{immune} (the predictor never sees the instruction), whereas designs that condition the \emph{predictor} on language (FiLM-into-the-predictor, e.g.\ \citealp{thinkjepa}) are exposed. We were unable to probe a \emph{released} external model directly: we found no public-weights world model that conditions its predictor on language, so our external evidence is the standard recipe trained on external \emph{environments} (BabyAI, Language-Table) rather than a published pretrained model. Probing such a model is the one remaining step to turn this characterization into a field-wide audit (\S\ref{sec:limits}).

\paragraph{Summary.} Across our studies on tabletop, BabyAI, and Language-Table, the leak is present exactly when the instruction names the scored quantity, and the dose-response shows it does not depend on non-instruction predictor strength. Because the model transcribes even against a perfect action, the remedy must remove the goal from the dynamics (\S\ref{sec:fix}); better features cannot prevent it.

\section{Discussion}
The central lesson is methodological and transferable. When a goal-conditioned world model is evaluated on a relation whose name appears in the instruction, a high relation-readout is not evidence of grounding; it can be pure instruction transcription, and it contaminates control as well as representation. The inexpensive, decisive controls are to \emph{withhold} the goal and to \emph{substitute} a counterfactual one, on the training render distribution. The architectural remedy is equally simple and is, in fact, the textbook planning setup: the goal belongs to the cost, not the transition. With the goal removed from the dynamics and the reference read path supervised, explicit anchors do ground relations genuinely, up to a distractor threshold, which is the fix's payoff. The control story is more sober: goal-conditioning monotonically hurts, and removing it merely \emph{recovers} control to the simplest no-goal baseline rather than improving on it, so we claim grounded perception, not a control advantage.

Two boundaries temper the positive result and are, we think, the more interesting science. The grounding is sharply bounded in ambiguity and does not transfer across ambiguity levels, which suggests the supervised anchors latch onto distribution-specific cues rather than a general relational percept. Moreover, once leakage is removed, the bottleneck for control is no longer the representation but the \emph{fidelity} of the learned dynamics: a one-step predictor whose multi-step rollouts drift cannot plan as well as the true dynamics even when its one-step anchors are accurate. These point to concrete next steps (multi-step prediction, action calibration, mixed-ambiguity training) rather than to the representation.

\section{Limitations}
\label{sec:limits}
Our claims are bounded by the following, which we state plainly. \emph{(i) Seed and scale coverage is partial.} Due to limited computational resources, the leakage collapse ($0.90\!\to\!0.27$) and the control gradient are run at three training seeds (mean$\pm$s.d.; orderings hold in every seed, \S\ref{sec:fix}), but the counterfactual probe ($94.5\%$ and $2.3\%$, $N{=}256$) and the across-ambiguity curves (Fig.~\ref{fig:perception}) are \emph{single-seed}; we do not run a full $\text{seed}\times\text{ambiguity}$ grid. The seed replication is at the medium-ambiguity operating point (amb2). \emph{(ii) Synthetic 2D task; no 3D or real-world validation.} The relational tabletop is controlled and toy. Constrained by compute budget, we validate the leakage phenomenon on two external \emph{environments} (BabyAI, Language-Table) but not on 3D scenes, real images, or high-dimensional action spaces. Scaling to such settings would strengthen the generality claim but requires substantially more training and data. \emph{(iii) No external pretrained model probed.} We could not find a public-weights world model that conditions its predictor on language; our external evidence therefore uses the standard recipe trained on external environments rather than probing a published pretrained checkpoint. Demonstrating the leak in a released model is the key remaining step. \emph{(iv) Dataset property.} Our instructions name the target relation; this is what enables the leakage we study, and although our fix (goal-free dynamics) is leakage-free \emph{by construction} regardless of the dataset, a stronger design would decorrelate the goal from the current relation. \emph{(v) Control: a tie, not a win, and below oracle.} The de-leaked model's control \emph{ties} the simplest no-goal baseline (\anchors) rather than beating it (the fix's value is grounded perception, not a control gain), and all variants remain well short of the oracle, which we attribute to dynamics fidelity. A multi-step prediction objective and action calibration, which we leave to future work due to resource constraints, would likely narrow this gap. \emph{(vi) Learned-anchor metric.} Our geometric readout assumes the supervised slot convention and is therefore not a fair test of \emph{learned} (unsupervised) anchors; we do not report a learned-versus-supervised verdict.

\section{Conclusion}
On a controlled relational task, a compact world model with explicit references appears to ground spatial relations but is, under goal conditioning, transcribing the instruction: a shortcut that inflates representation metrics and degrades control, and that is invisible without a goal-withheld or counterfactual control. Moving the goal out of the dynamics, where it does not belong, recovers genuine instruction-independent grounding (the fix's payoff) and, for control, shows that goal-conditioning was the culprit: removing it restores control to the simplest no-goal baseline, with the residual gap to an oracle attributable to dynamics fidelity rather than the representation. The diagnosis, the detection protocol, and the one-line architectural fix apply to goal-conditioned world models whose instruction names the scored quantity. A survey of recent designs (\S\ref{sec:external}) finds that the split between exposed and immune designs is structural: designs that place the goal in the planner's \emph{cost} (e.g.\ \citealp{vlwm,vlmpc,pijepa}) are immune by construction, while designs that condition the \emph{predictor} on language (e.g.\ FiLM-into-the-predictor \citealp{thinkjepa}) are exposed. We could not probe a \emph{released} external model directly, as no public-weights world model conditions its predictor on language, so demonstrating the leak in a published pretrained model (rather than the standard recipe trained on external \emph{environments}, as here) remains the key step to convert our characterization into a field-wide audit.

\bibliographystyle{tmlr}
\bibliography{refs}

\end{document}